\documentclass[letterpaper, 10 pt, conference]{ieeeconf}  
\usepackage{graphicx}
\usepackage{amsmath}
\usepackage{amssymb}
\usepackage{booktabs}
\usepackage{multirow}
\usepackage{hyperref}
\usepackage[table]{xcolor}

\definecolor{goodgreen}{RGB}{0,120,0}
\definecolor{badred}{RGB}{180,0,0}
\definecolor{lightgreen}{RGB}{220,245,220}

\IEEEoverridecommandlockouts                              

\title{\LARGE \bf
Abstention-Aware Personalized Object Rearrangement via Uncertainty-Guided LLM Assistance
}

\author{Sam Collin$^{1*}$ and Ali Ayub$^{1}$
\thanks{
This research was undertaken, in part, thanks to funding from the Natural Sciences and Engineering Research Council of Canada (NSERC), Concordia University, 
and computing resources from the Digital Research Alliance of Canada.}
\thanks{$^{1}$Concordia University, Montreal, Quebec H3G 1M8, Canada}
        \thanks{{\tt\small $*$sam.collin@mail.concordia.ca, ali.ayub@concordia.ca}}%
}

\begin{document}

\maketitle
\thispagestyle{empty}
\pagestyle{empty}

\begin{abstract}

Robotic assistance in household environments requires not only predicting where objects should be placed, but also reasoning about when objects should not be placed at all. Existing approaches to personalized object rearrangement primarily focus on placement decisions under the assumption of clean observations and complete actionability, limiting their applicability in realistic, cluttered, and partially erroneous settings. In this paper, we introduce APOLLO, a hybrid framework for abstention-aware personalized object rearrangement that combines a lightweight, personalized embedding model (PEM) with selective large language model (LLM) assistance. PEM is trained for each user–environment pair using a small number of demonstrations, operates entirely on CPU, and produces uncertainty estimates, which are used to selectively invoke LLM-based reasoning only for ambiguous decisions, balancing efficiency, privacy, and reasoning capability. To evaluate this formulation beyond existing benchmarks, we introduce APOR, a synthetic, LLM-generated dataset that captures room-level, multi-furniture environments, diverse organizational profiles, explicit abstention behavior, and noisy partial scene context. Extensive experiments on both PARSEC and APOR provide initial evidence that APOLLO improves over prior LLM-based baselines in controlled benchmark settings while substantially reducing LLM usage. Code is available at https://github.com/PaInt-Lab/APOLLO.

\end{abstract}


\section{Introduction}

\noindent
Coming home to a messy environment is a familiar experience and can be frustrating (e.g. a mug left on the worktop, a vacuum cleaner left lying around). However, what appear as clutter may seem perfectly normal from another perspective. 
For a robot, tidying up a space is not a simple pick-and-place task but an ongoing process of understanding the environment and the user to restore it to a desired state~\cite{ayub2024interactive}. Following Batra et al.~\cite{batra_rearrangement_2020}, we view rearrangement as the process of transforming an environment to reach a target state through appropriate object placements. 
What makes the situation so difficult here is not only the combinatorial nature of the scene, but above all, that the `right’ arrangement is inherently user-specific. The rules or habits of a household are not written down; they develop over time, potentially through conflicts between users and the environment. While certain general patterns may be found among some users, it is common to find individual habits or quirks that defy common sense. For example, a user may follow a strategy based on visibility, because an object that is not visible to them is likely to be forgotten, or another user may organize their belongings emotionally, leaving the things they love on display and hiding the rest. These preferences become increasingly difficult for the robot to understand as environments become more complex (multiple pieces of furniture, more potential surfaces, cluttered partial states).

\begin{figure}[t]
\centering
\includegraphics[width=0.8\columnwidth]{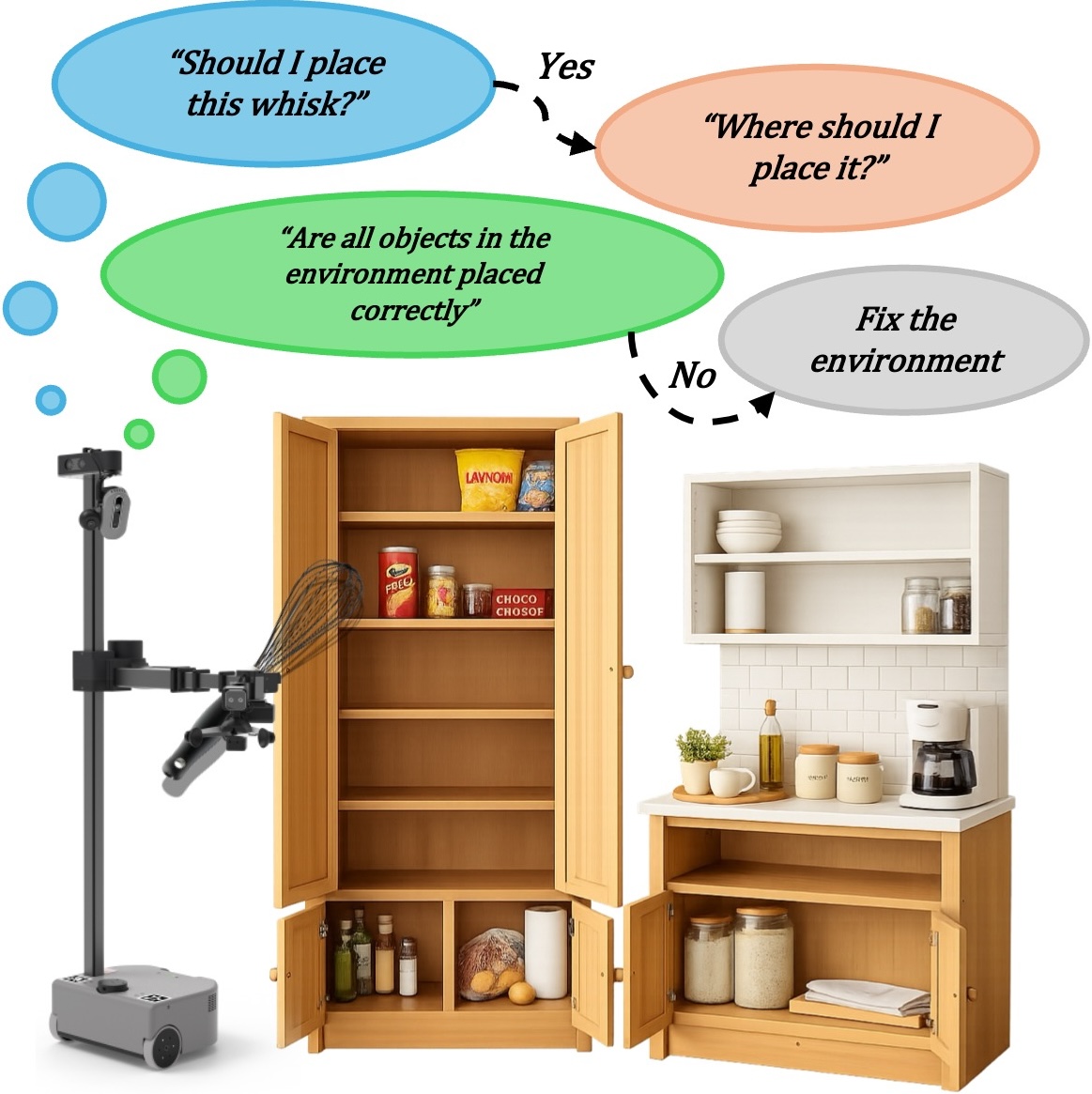}
\caption{\textbf{Overview.} 
Given an environment, the robot must first observe and determine if any objects in the environment must be correctly placed, and then fix them based on user preferences. Given a new object, the robot first needs to decide if it should be placed in the environment, and then determine the appropriate surface to place the object based on user preferences.}
\label{fig:intro_diagram}
\end{figure} 

Previous studies have approached the task of personalized rearrangement from several different angles. Traditional methods are based on generic common sense \cite{sarch_tidee_2022} derived from large datasets and often require offline training, which prevents user-specific adaptation by over-generalizing \cite{abdo_robot_2015, kapelyukh_my_2021}. In personalized rearrangement, several methods use either prior context or in-scene context alone. However, relying only on prior context may differ from the actual state of the environment \cite{newman_degustabot_2024,wu_tidybot_2023}, and context-only methods may struggle with sparsely populated scenes \cite{ramrakhya_seeing_2024,ramachandruni_consor_2023}. While some works combine the two signals~\cite{ramachandruni_personalized_2025}, they either assume the in-scene context is entirely reliable~\cite{ramachandruni_personalized_2025} or require additional interactions to clarify preferences and manage constraints~\cite{wang_apricot_2024}, which might not be feasible in real-world environments. Furthermore, these methods often rely heavily on LLMs/VLMs to improve few-shot preference inference but raise concerns about latency, security, and cost \cite{wang_apricot_2024,ramachandruni_personalized_2025,han_llm-personalize_2025}. Finally, prior work also assumes that all the objects available to the robot must be placed on a given furniture item, which is likely unrealistic, and the robot might need to abstain from placing objects in the wrong place.

To address these limitations, we present Abstention-aware Personalized Object Rearrangement with uncertainty-guided LLM assistance (APOLLO), a hybrid 
framework that uses a lightweight model for learning personalized embeddings (PEM) to model user object placement preferences, and is assisted by an LLM when PEM outputs are uncertain. The model can learn from limited data, run on a CPU, tolerate noisy partial context, and can abstain from placing objects in the environment when appropriate. 
We also introduce Abstention-aware Personalized Object Rearrangement (APOR), a synthetic, LLM-generated benchmark designed to study complex multi-furniture environments and diverse organizational profiles in a controlled setting. It also embeds two complementary subtasks: \emph{placement}, i.e., deciding \emph{where to place} an object, and \emph{abstention}, i.e., deciding \emph{when not to place} an object to respect user preferences.
Extensive evaluations on a standard object placement benchmark, PARSEC~\cite{ramachandruni_personalized_2025}, and APOR, provide initial evidence that our method improves over prior LLM-based state-of-the-art methods for a fraction of the cost. Our results further suggest that existing models are not designed to capture abstention and struggle as environmental complexity increases, and when partially arranged scenes contain errors.


\section{Related Work}
\label{sec:related_work}
\noindent
Household rearrangement has been studied under a range of task formulations, from goal-driven scene restoration to commonsense tidying and personalized organization. Early rearrangement work formulates the task as reaching a specified target state under physical and semantic constraints, often in embodied simulation, where success is defined by whether objects are moved to predefined goal locations~\cite{batra_rearrangement_2020,goyal_ifor_2022}. Related tidying benchmarks relax the requirement for explicit targets and instead ask agents to identify misplaced objects and restore order using commonsense notions of organization~\cite{sun2025robotidy3dgaussian}. While these settings differ in how the desired arrangement is specified, they largely focus on deciding \emph{where} objects should be placed, assuming that all candidate objects belong in the environment and that the desired final arrangement is well defined.

A more closely related line of work studies \emph{personalized} rearrangement, where object placement depends on user-specific preferences inferred from prior interactions or demonstrations. Earlier approaches model such preferences through collaborative filtering or latent representations learned offline and then reused at inference time~\cite{kapelyukh_my_2021}. Extending this direction to partial observations, ConSOR predicts placements from in-scene contextual cues in partially arranged scenes~\cite{ramachandruni_consor_2023}. More recent systems incorporate LLMs to improve generalization from sparse demonstrations and partial scene context: PARSEC formalizes personalized rearrangement from demonstrations and partial context~\cite{ramachandruni_personalized_2025}; TidyBot uses LLM-based reasoning for personalized robotic assistance~\cite{wu_tidybot_2023}; and APRICOT combines prior preference information with scene context while handling ambiguity through user queries and constraints~\cite{wang_apricot_2024}. Despite strong performance, these methods generally focus on predicting a placement destination for each object, rather than deciding whether an object should be placed at all. In addition, methods that rely on cloud-hosted LLM inference can raise practical concerns regarding latency, cost, and privacy in real-world deployments~\cite{wu_tidybot_2023,wang_apricot_2024}.

A separate but conceptually related direction studies abstention or selective computation, where a system decides whether to act immediately or defer additional reasoning or computation~\cite{izzo_act_2026}. However, this notion of abstention differs from the setting considered in our work: here, abstention is an \emph{object-level rearrangement outcome}, corresponding to the decision that an object should remain unplaced because it is irrelevant to the user’s intended organization of the environment. Moreover, prior rearrangement methods typically assume that the observed environment is otherwise correctly arranged, whereas real household scenes may already contain placement errors. In contrast, our work introduces a new dataset and a personalized rearrangement framework that operates under limited compute, considers incorrectly arranged scenes, and selectively defers uncertain decisions to an LLM while abstaining from placing irrelevant objects.

\begin{figure*}
    \centering
    \includegraphics[width=0.9\linewidth]{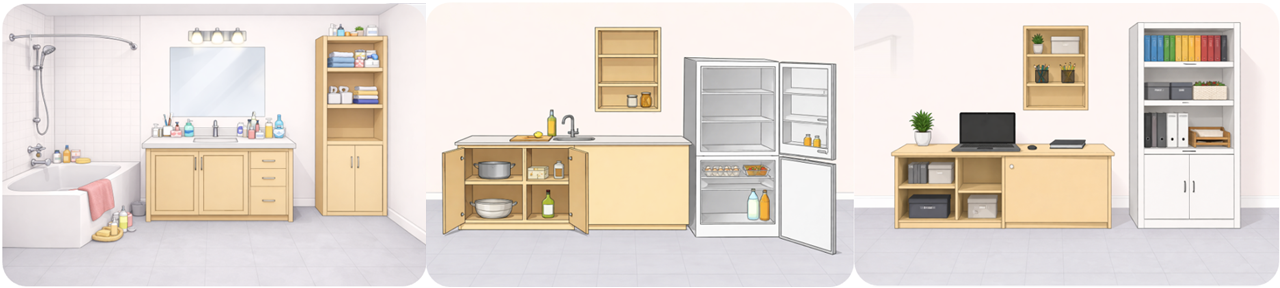}
    \caption{Examples of object arrangements in different rooms in APOR for three user profiles. (Left) \textit{Cluttercore} focuses on dense, layered placement, maximizing visible objects and aesthetic richness. (Middle) \textit{Low Reach} places objects within reachable low and mid-level surfaces. (Right) \textit{Taxonomic} groups objects strictly by category, regardless of context or use.}
    \label{fig:profile_examples}
\end{figure*}

\section{Problem Formulation}
\noindent
We formalize the object rearrangement task 
as an extension of the formulation in previous work~\cite{ramachandruni_personalized_2025}, adding multiple furniture items and an explicit abstention outcome. We index each setting by a \emph{user--environment} pair $(u,e)$, where $u$ denotes the user and $e$ the environment.
We assume that each environment $e$ consists of a set of surfaces $\mathcal{S}$ (on multiple furniture items) on which objects from a closed vocabulary $x \in \mathcal{O}$ can be placed to form a specific arrangement, defined as an object-surface pair: $\mathcal{A} = \left\{(x,s)\right\}$. For each evaluation instance, the target arrangement is denoted as the ground truth $\mathcal{A}^*_G$. Unlike previous work, the target arrangements in our benchmark have objects that are deliberately left unplaced. 
Therefore, we consider a modified set of possible surfaces $\mathcal{S}^+ =$ $\mathcal{S}$ $\cup$ \{\texttt{UNPLACED}\}. The role of the \texttt{UNPLACED} surface is to measure abstention by allowing the model to explicitly refrain from placing an object due to a learned preference. We denote by $\mathcal{X}$ the set of objects present in a target scene $\mathcal{A}^*_G$, with $\mathcal{X} \subseteq \mathcal{O}$. 

At inference time, the agent must propose a transformation that enables a specific goal state $\mathcal{A}^*_G$ to be obtained from an arrangement in a partial state $\mathcal{A}_P$. The partial arrangement $\mathcal{A}_P$ is formed by moving a subset of objects $\mathcal{X}_U \subseteq \mathcal{X}$ from the goal arrangement to the \texttt{UNPLACED} surface. In some evaluation settings, $\mathcal{A}_P$ may also contain deliberately corrupted in-scene placements to test sensitivity to noisy partial context.
The number of objects moved from $\mathcal{X}$ to $\mathcal{X}_U$ varies between instances, satisfying $1 \leq |\mathcal{X}_U|\leq |\mathcal{X}|$. The agent can then plan and act on this set of ``actionable'' objects, by deciding on a placement or leaving the object unplaced to propose a predicted set $\mathcal{A}_G$ comparable to $\mathcal{A}^*_G$.

To guide its prediction and best match the user's preferences, the agent can rely on two different types of signals. The \textit{prior-scene} context, called user history $\mathcal{H}_{u,e} = \left\{A_H^1, A_H^2,...,A_H^N\right\}$, provides the model with examples of scenes where user $u$ has placed other pools of objects within the same environment $e$. 
For additional information, the agent can also rely on the placement of objects already in the partially arranged scene $\mathcal{A}_P$ at the time of prediction, the \textit{in-scene} context $\mathcal{X}_P \subseteq \mathcal{X}$, ranging from 0 (empty partial) up to $|\mathcal{X}|-1$. This provides an additional signal about the user and the configuration in which they wish to organise their space. These two signals are highly complementary, as conflicting preferences inferred from history can be resolved by elements already present.

\section{Dataset}\label{sec:dataset}
\noindent We introduce \textit{Abstention-aware Personalized Object Rearrangment (APOR)}, a synthetic benchmark for personalized object rearrangement. Our goal is to study 
two axes that have not been addressed in previous datasets: (i) atypical user preferences (non-standard but consistent organizational logic), and (ii) more complex 
environments (room-level, noisy partial context, objects that must not be placed).  As large amounts of this specific type of data are difficult to collect, we generated APOR with an LLM-assisted pipeline and lightweight constraints (validation and retries). These constraints enforce structural validity of the generated scenes, but do not constitute human validation of the underlying preferences.
The pipeline is modular and can be extended with new profiles, environments, and object pools. Table~\ref{tab:apor_summary} summarizes the resulting benchmark and evaluation split.

\begin{table}[t]
\centering
\footnotesize
\setlength{\tabcolsep}{4pt}
\renewcommand{\arraystretch}{1.05}
\caption{\textbf{APOR dataset summary.} Top: global statistics. Bottom: evaluation profiles with the number of \emph{clean} scenarios (excluding error modes) and Ground Truth (GT) abstention rate (fraction of actionable objects labeled \texttt{UNPLACED}).
}
\label{tab:apor_summary}
\begin{tabular}{l}
\toprule
\multicolumn{1}{c}{\textbf{Global statistics}}\\
\midrule
\multicolumn{1}{c}{
\begin{tabular}{l r @{\hspace{12pt}} l r}
Profiles & 10 & Room Types & 5 \\
Users per profile & 3 & Users Total & 30 \\
Furniture Total & 60 & Furniture per env & 3--6 \\
Environments & 150 & Arrangements per $(u,e)$ & 6 \\
Arrangements Total & 900 & Env. complexity score & $[0,1]$ \\
Object Vocabulary & 581 & Objects per arrangement & Up to 20 \\
Scenarios (w/out errs) & 3792 & Overall GT-unplaced ratio & 0.17 \\
Scenarios (w/ errs) & 1992 & Total Scenarios & 5784\\
\end{tabular}
}\\
\midrule
\multicolumn{1}{c}{\textbf{Profiles}}\\
\midrule
\multicolumn{1}{c}{
{\setlength{\tabcolsep}{2.5pt}
\begin{tabular}{l r r @{\hspace{12pt}} l r r}
\textbf{Profile} & \textbf{\#Scen.} & \textbf{unp\_ratio} & \textbf{Profile} & \textbf{\#Scen.} & \textbf{unp\_ratio}\\
\midrule
Room & 345 & 0.41 & Thematic & 391 & 0.27 \\
Frequency & 418 & 0.19 & Ladybug & 420 & 0.21 \\
Exclusionist & 228 & 0.72 & LowReach & 400 & 0.25 \\
Cluttercore & 450 & 0.03 & HighReach & 399 & 0.25 \\
Taxonomic & 356 & 0.35 & Efficient & 385 & 0.30 \\
\end{tabular}
}}\\
\bottomrule
\end{tabular}
\end{table}

\paragraph*{\textbf{Environments}}
We have targeted five types of household environments: kitchen, bathroom, living room, bedroom, and office. 
Each environment contains 3-6 pieces of furniture sampled from a catalog of 21 different items. There are up to three variations per furniture type, totaling 60 different pieces, which increase structural diversity. 
Each piece is defined by a description and a list of available surfaces. All attributes have been deliberately kept neutral to avoid introducing any functional bias (e.g. no "vegetable drawer"). To ensure consistency, we ensure that certain items of furniture are present in the different types of room (e.g., a shower or bathtub in the bathroom). We also assign a per-environment $[0,1]$ score based on the structural complexity of the furniture present in the scene.

\paragraph*{\textbf{Pool of objects}}
APOR contains a collection of 581 pickable objects originating from two sources: 62 from AI2-THOR (iTHOR scenes)~\cite{kolve_ai2-thor_2022} and 519 LLM-generated objects, added to cover the specific needs of each profile. Objects are defined by their unique name and descriptive attributes (e.g., color, room hints, and associated profiles). For each scene, 
approximately half of the object pool is room-relevant according to catalog room hints, and the rest are sampled from the full catalog, with at most two instances of the same canonical object and a maximum of 20 objects per arrangement. 

\paragraph*{\textbf{User Profiles}}
Profiles are archetypes that describe coherent but potentially non-standard organizational strategies. In this paper, our evaluation is based on the 10 profiles listed in Table~\ref{tab:apor_summary}. Inspiration for these profiles comes from previous crowdsourcing~\cite{ramachandruni_personalized_2025} and behavioral studies~\cite{rowe_desk_2019, mirman_taxonomic_2017} and are further refined into self-consistent, LLM-generated descriptions (traits and organizing principles). Fig~\ref{fig:profile_examples} shows examples of object arrangements generated in APOR for three different user profiles. Other profiles introduced in the benchmarks are context-driven (Room), theme-based (thematic), usage-driven (Frequency), repetitive, pattern-based (Ladybug), selective (Exclusionist), reach-constrained (LowReach), density-driven (Cluttercore), height-driven (HighReach), category-driven (Taxonomic), and efficiency-driven (Efficient) organizational profiles.

\begin{figure*}[t]
\centering
\includegraphics[width=.9\textwidth]{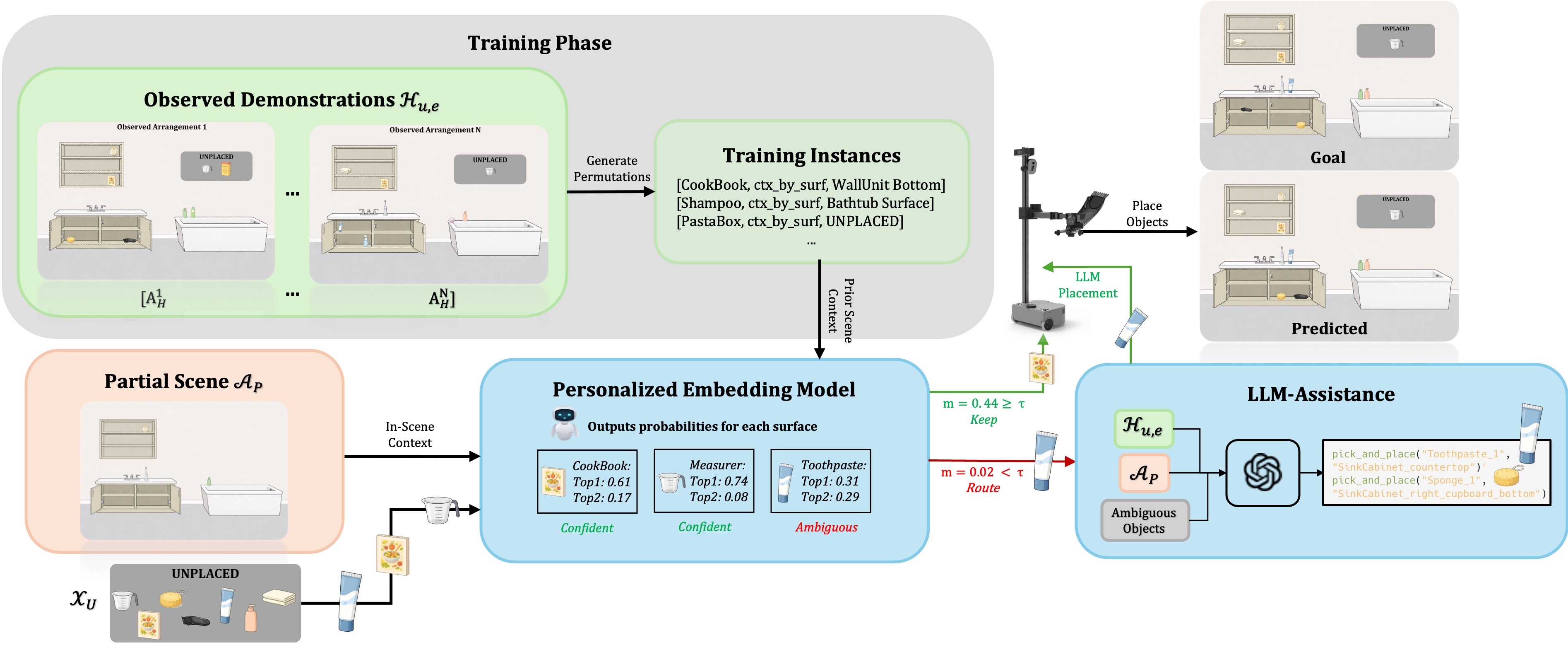}
\caption{Overview of the APOLLO framework. PEM is trained per $(u,e)$ from observed demonstrations via permuted training instances, then scores candidate surfaces for each actionable object in a partial scene to produce per-object probabilities and confidence (margin). APOLLO keeps confident local decisions and routes only ambiguous objects to an LLM, merging both outputs into the final predicted arrangement sent to the robot for manipulation.}
\label{fig:apollo_diagram}
\end{figure*}

\paragraph*{\textbf{Data Generation}}

In the base dataset, we generated the layouts using a locally hosted LLM (Gemma-27B \cite{team_gemma_2025}), chosen to keep generation local and reproducible. 
The model is tasked to embody one of the profiles and to produce object-to-surface placements in the current environment. This choice is motivated by previous research exploring LLMs as simulators of human-like behavior~\cite{zhang_large_2023,park_generative_2023}, but APOR should not be interpreted as validated human preference data. 
To preserve continuity within the room between arrangements of the same $(u,e)$ couple, we provide up to three previous arrangements as context.
In total, there are \textbf{10} preference archetypes, each represented by \textbf{3} synthetic users (30 total). For each of these users, we generated \textbf{5} room-level environments (one per room-type) and sampled \textbf{6} arrangements per environment. This represents a total of \textbf{150} environments and \textbf{900} $(u,e,\mathcal{X})$ scenes.

\paragraph*{\textbf{Controlled Abstention}}\label{para:local}
To ensure control over the number of objects originally left unplaced by the model, which varies greatly depending on the profile, we have added a post-processing step. For each arrangement, and at random, only \textbf{40\%} of the objects unplaced by the model are retained. This transformation preserves the same number of users and arrangements, while reducing the number of unplaced objects for a more balanced overall abstention rate (\textbf{0.17}). As a result, the abstention distribution in APOR is partially design-controlled rather than fully emergent from the LLM-generated profiles.


\section{Methodology}
\noindent

\noindent
Fig~\ref{fig:apollo_diagram} provides an overview of
the APOLLO framework, which relies primarily on a lightweight scorer to make rapid object-level decisions and queries the LLM when uncertain about its decisions.

\subsection{PEM: Personalized Embedding Model}\label{sec:pem}

The Personalized Embedding Model (PEM), is a CPU-only embedding-based model that naturally captures co-occurrence patterns. It is trained from scratch for each $(u,e)$ couple using only the user-history $\mathcal{H}_{u,e}$ (no cross-user pretraining). Similar to prior work for personalization using personal embeddings~\cite{paleja2020interpretable, schrum2022mindmeld}, PEM learns personalized embeddings for objects and surfaces.
Given an actionable object $x \in \mathcal{X}_U$, the model predicts a candidate surface $s \in \mathcal{S}^+$ and outputs a confidence signal used later for LLM assistance.

\paragraph*{Inputs} For each new scene, the model receives the following inputs: (i) observed full arrangements from the same ($u$,$e$) couple. (ii) the partial scene $\mathcal{A}_P$ which contains object-surface pairs that are already placed in the scene (the partial scene might have some errors). (iii) the list of actionable objects, i.e., objects on the \{\texttt{UNPLACED}\} surface that require an action from the model.

\paragraph*{Training Instances} We generate supervised training instances from $\mathcal{H}_{(u,e)}$ by constructing partial scenes through object removals. 
For each demonstration $\mathcal{A}_H^{i} \in \mathcal{H}_{(u,e)}$, we first extract the set of placed object-surface pairs $\{(x,s)\}$. 
We then generate up to $C_{max}$ synthetic partial scenes from that demonstration, where this cap is used to avoid imbalance across arrangements. For each synthetic partial scene, we sample the number of removed objects as $r \sim \mathcal{U}\{1, \ldots, \min(r_{\max}, |\mathcal{X}_H^{i}|)\}$, and reassign them to {\texttt{UNPLACED}\}, while leaving all remaining placed objects unchanged.
Each resulting partial scene is encoded as a context-by-surface structure (ctx-by-surface) representation of length $|\mathcal{S}^+|$. Each entry stores the indices of objects currently present on that surface in the partial scene. From this encoding, we generate training instances of the form 
$$
(\textit{obj\_idx},\ \textit{ctx\_by\_surface},\ \textit{target\_surface\_idx}).
$$

This representation allows the model to learn an object’s target surface conditioned on the current scene state.

\paragraph*{Representations}
PEM learns object and surface embeddings of dimension $d$. We denote $E_o(x) \in \mathbb{R}^d$ as the embedding of object $x$ and $E_s(s) \in \mathbb{R}^d$ as the embedding of surface $s$. Each surface also has a key vector $K(s) \in \mathbb{R}^d$ used to gate the global context to make it surface-dependent, a learned surface bias $b_s \in \mathbb{R}$ and a learned token $e_{\emptyset} \in \mathbb{R}^d$ to represent empty surfaces. All parameters learned for a given $(u,e)$ are:
$$
\theta = \{E_o, E_s, K, b, e_\emptyset\}
$$

\paragraph*{Local and global context encoding}
PEM encodes the partial scene at two complementary scales. First, it builds a \textit{surface-level} summary $c_s$ to capture local grouping cues (what is already placed together on a given surface). Second, it builds a compact \textit{scene-level} summary $g$ to capture broader organization patterns. Let $C_s$ be the set of objects present on surface $s$ in the partial context. The surface-level context is the average of their object embeddings:
$$
c_s =
\begin{cases}
\frac{1}{|C_s|} \sum_{x \in C_s} E_o(x) & \text{if } |C_s| > 0 \\
0 & \text{otherwise}
\end{cases}
\in \mathbb{R}^d
$$

When forming the global scene summary, we want the empty surfaces to still contribute information ("surface currently unused") rather than being indistinguishable from missing data. The resulting scene-level summary is then computed as the average of these per-surface context vectors ($c_s$ for non-empty surfaces, $e_{\emptyset}$ for empty ones):
$$
g = \frac{1}{|S^+|} \sum_{s \in S^+} \left(c_s + \mathbf{1}[|C_s| = 0] \cdot e_{\emptyset}\right) \in \mathbb{R}^d
$$
A single global context applied uniformly to all surfaces is restrictive. We therefore assign each surface a learned key vector $K(s)$ and compute a surface-dependent gate as:
$$
\alpha_s = \sigma\!\left(K(s)^{\top} g\right) \in (0,1).
$$
The sigmoid maps the compatibility between the surface and the global context to a weight in $(0,1)$, making the influence of $g$ surface-dependent: it lets the model emphasize global context when a surface needs disambiguation, and downweight it when local grouping cues are already sufficient.

Finally, we denote $v_s$ as the surface representation used for scoring, obtained by combining the surface embedding with the local context and the gated global context contribution:
$$
v_s = E_s(s) + c_s + \alpha_s g \in \mathbb{R}^d
$$

\paragraph*{Scoring and training}
Given an actionable object $x \in \mathcal{X}_U$, we compute logits over all candidates surfaces:
$$
score(x, s) = E_o(x)^{\top} v_s + b_s, \quad s \in S^+.
$$
A softmax is applied to obtain $p_{\theta}(s \mid x)$ and we train with cross-entropy loss against the target surface. Parameters $\theta$ are optimized using Adam for a fixed number of epochs.

\paragraph*{Inference}
At test time, we construct the surface representation $v_s$ from the objects already placed in the partial scene $\mathcal{A}_P$, calculate the probability distribution $p_\theta(s \mid x)$ for each of the actionable objects, and predict:
$$
\hat{s}_{PEM}(x) = \arg\max_{s \in S^+} p_\theta(s \mid x)
$$

\subsection{LLM-assisted inference}\label{sec:ose}
PEM is designed to capture user- and environment-specific placement patterns, but it may benefit from the reasoning capabilities of LLMs when prediction uncertainty is high. To decide \emph{when} to request assistance, we use a simple uncertainty signal derived from PEM's own output distribution. Specifically, we define the \emph{margin} $m(x) = p_1(x) - p_2(x)$, where $p_1(x)$ and $p_2(x)$ are the top-1 and top-2 probabilities predicted by PEM for object $x$. A small margin indicates that multiple surfaces are nearly tied, suggesting uncertainty. For a given threshold $\tau$, we query the LLM for object $x$ if $m(x) < \tau$:
$$
\small
\hat{s}_{APOLLO}(x) =
\begin{cases}
\arg\max_{s \in \mathcal{S}^+} \; p_{PEM}(s \mid x) & \text{if } m(x) \ge \tau \\
\hat{s}_{LLM}(x) & \text{otherwise}
\end{cases}
$$
The final prediction $\hat{\mathcal{A}}_G$ is obtained by merging all per-object decisions $\hat{s}(x)$ with the objects already placed in $\mathcal{A}_P$.


\section{Experiments}
\label{sec:experiments}

\noindent
We compare our method against previous baselines on PARSEC~\cite{wu_tidybot_2023, ramachandruni_personalized_2025, wang_apricot_2024} and our APOR benchmark (Sec~\ref{sec:dataset}) in a controlled benchmark settings to evaluate the model's ability to deal with unaddressed challenges in prior work.

\subsection*{Benchmarks and evaluation protocol}
\textbf{PARSEC} is a placement-only benchmark that evaluates object-to-surface placement without abstention \cite{ramachandruni_personalized_2025}. 
Its evaluation phase comprises 3,507 scenarios representing 72 users in 15 different single-furniture environments, reported over the \textit{NovelEnvCat} split. In PARSEC, each $(u,e)$ provides six arrangements; iteratively, one is selected as the target $\mathcal{A}^*_G$ and the remaining five form the history $\mathcal{H}_{u,e}$. The actionable set $\mathcal{X}_U$ is constructed using all possible partial sizes.

\textbf{APOR:} We evaluate on the APOR benchmark presented in Section~\ref{sec:dataset}, using the \texttt{reduced40} variant. As in PARSEC, each $(u,e)$ provides six arrangements with one selected as the target $\mathcal{A}^*_G$ and the other five used for history $\mathcal{H}_{u,e}$. The difference lies in how we construct evaluation scenarios from each target arrangement. 
Instead of enumerating all possible partial sizes, we sample a small set of modes when feasible: \textit{empty} where the partial scene is empty ($\mathcal{A}_P=\emptyset$), \textit{missing1} where exactly one object is moved from the goal to \texttt{UNPLACED}, and \textit{placed$k$\_correct} with $k\in\{5,10,15\}$, where $k$ objects remain correctly placed in $\mathcal{A}_P$ and the remaining actionable objects are moved to \texttt{UNPLACED}.
We additionally create \textit{placed$k$\_err$e$} modes with $(k,e)\in\{(5,1),(10,2),(15,3)\}$, where $e$ of the $k$ in-scene objects are deliberately placed on incorrect surfaces to simulate noisy in-scene context. Importantly, the goal $\mathcal{A}^*_G$ is unchanged, and we do not score these injected errors directly; they only affect the model through the (possibly corrupted) input context, and we report their effect separately in Table~\ref{tab:pa_mix_k5_k10_errors}.

\subsection*{Baselines} 
We compare APOLLO against LLM-based baselines from prior work, including \textbf{ContextSortLM} (CSLM) \cite{ramachandruni_personalized_2025}, \textbf{TidyBot} \cite{wu_tidybot_2023} and \textbf{APRICOT} \cite{wang_apricot_2024} on the PARSEC dataset. Following the PARSEC setup, TidyBot-Random is a variant that randomly samples a demonstration from the history $\mathcal{H}_{u,e}$ to comply with TidyBot’s single-example rule induction, and APRICOT-NI for Non-Interactive constructs a single textual preference description from $\mathcal{H}_{u,e}$ without interactive questioning. For APOR experiments, we compare against CSLM only. For fairness, CSLM is evaluated with an action space that includes \{\texttt{UNPLACED}\}. For both PARSEC and APOR experiments, we also report PEM-only as an ablation of APOLLO, isolating the contribution of selective LLM assistance.

\subsection*{Implementation details} 

\textbf{APOLLO (PEM + CSLM assistance):} We run APOLLO as a two-stage pipeline in which PEM outputs per-object placement probabilities (and the margin $m(x)=p_1(x)-p_2(x)$), and the LLM assistant uses the same prompt structure as CSLM. 
We use embedding dimension $d=32$ and train PEM for $\mathcal{E}=60$ epochs (batch size $B=16$, $lr=0.01$, weight decay $\lambda=0.001$). Training instances are generated from demonstrations by sampling up to $C_{\max}=8$ synthetic partial scenes per observed arrangement, removing up to $r_{\max}=6$ objects (moved to \texttt{UNPLACED}). By design, PEM runs entirely on CPU; all PEM experiments were run on an Apple M4 Pro. At inference time, APOLLO uses the LLM for an object $x$ when $m(x)<\tau$. Thresholds were chosen from an offline sweep over fixed PEM and CSLM predictions: $\tau=0.15$ on PARSEC and $\tau=0.05$ on APOR. We report PEM and APOLLO results over five PEM seeds (42--46), with CSLM predictions fixed. Runtime estimates use PEM forward prediction time, full CSLM reasoning time, and the observed fraction of objects routed by APOLLO.

\textbf{LLM-Based Models:} On PARSEC, we reuse the original predictions obtained under the evaluation protocol (OpenAI API; GPT-4 model) to ensure strict comparability with reported baselines. For APOR, all CSLM-based calls, including APOLLO's LLM fallback, use the same locally hosted Llama-3.3-70B-Instruct setup~\cite{grattafiori_llama_2024}: 2$\times$H100 GPUs, bfloat16 precision, automatic device mapping, max\_new\_tokens=2048, temperature 1/top-p 1 for summary and placement prompts, and temperature 0/top-p 1 for rule extraction.

\textbf{Runtime estimation:} We report PEM forward prediction time and CSLM full prompt-based reasoning time per evaluated case. For APOLLO, effective runtime is estimated as
$t_{\mathrm{APOLLO}} = t_{\mathrm{PEM}} + \rho_{\mathrm{obj}} t_{\mathrm{CSLM}}$,
where $\rho_{\mathrm{obj}}$ is the fraction of actionable objects routed to the LLM.

\begin{table}[t]
\centering
\caption{Performance on novel environment categories on uniform (U) 1-D, 2-D, and non-uniform environments, along with the average, in the PARSEC dataset.}
\begin{tabular}{lcccc}
\toprule
\multirow{2}{*}{\textbf{Model}} &
\multicolumn{4}{c}{\textbf{NovelEnvCategory}} \\
\cmidrule(lr){2-5}
& U-1D & U-2D & Non-Uniform & \textbf{Average} \\
\midrule
APOLLO & \textbf{0.56} & \textbf{0.57} & \textbf{0.67} & \textbf{0.61} \\
PEM & \textbf{0.56} & \underline{0.54} & \underline{0.65} & \underline{0.59} \\
CSLM & \underline{0.54} & \textbf{0.57} & \underline{0.65} & \underline{0.59} \\
APRICOT-NI & 0.50 & 0.50 & 0.56 & 0.53 \\
TidyBot-Random & 0.46 & 0.40 & 0.46 & 0.44 \\
\bottomrule
\end{tabular}
\label{tab:tab2}
\end{table}

\begin{table*}[t]
\centering
\footnotesize
\setlength{\tabcolsep}{3pt}
\renewcommand{\arraystretch}{1.2}
\caption{\textbf{APOR results by user profile.} 
For each user profile, we report scene-level macro placement accuracy for all models. Each cell reports $\mathrm{pa}_{mix}$, except the last column, which reports the unweighted average across profiles as the triplet $\mathrm{pa}_{mix} \mid \mathrm{pa}_{place} \mid \mathrm{pa}_{abs}$.}

\begin{tabular}{lccccccccccc}
\toprule

\multirow{2}{*}{\textbf{Model}} &
\multicolumn{10}{c}{\textbf{Profile}} &
\multicolumn{1}{c}{\textbf{Average}} \\

\cmidrule(lr){2-11}
\cmidrule(lr){12-12}

& Room & Frequency & Exclusionist & Cluttercore & Taxonomic & Efficient & Thematic & Ladybug & LowReach & HighReach 
& {\scriptsize $\mathrm{pa}_{mix} \mid \mathrm{pa}_{place} \mid \mathrm{pa}_{abs}$} \\
\midrule

CSLM
& 0.25
& \underline{0.25}
& \underline{0.26}
& 0.20
& 0.25
& 0.27
& 0.31
& \underline{0.26}
& 0.22
& 0.27
& 0.25 $\mid$ \textbf{0.30} $\mid$ 0.22 \\

PEM 
& \textbf{0.53}
& 0.24
& \textbf{0.77}
& \underline{0.26}
& \textbf{0.44}
& \textbf{0.39}
& \underline{0.32}
& 0.25
& \textbf{0.31}
& \underline{0.30}
& \underline{0.38} $\mid$ 0.17 $\mid$ \textbf{0.56} \\

APOLLO 
& \underline{0.47}
& \textbf{0.29}
& \textbf{0.77}
& \textbf{0.28}
& \underline{0.41}
& \underline{0.38}
& \textbf{0.35}
& \textbf{0.30}
& \underline{0.29}
& \textbf{0.33}
& \textbf{0.39} $\mid$ \underline{0.25} $\mid$ \underline{0.47} \\

\bottomrule
\end{tabular}
\label{tab:tab3}
\end{table*}

\subsection*{Metrics}
To measure the models’ performance on the two subtasks: placement and abstention, we calculate the placement accuracy $pa$ in three variants: (i) $\mathbf{pa}_{\mathbf{mix}}$, computed by comparing the predicted arrangement $\mathcal{A}_G$ to the goal $\mathcal{A}^*_G$ over the full action space (including \texttt{UNPLACED}); (ii) $\mathbf{pa}_{\mathbf{place}}$, computed on the subset of actionable objects whose ground-truth surface is not \texttt{UNPLACED} (``where to place''); and (iii) $\mathbf{pa}_{\mathbf{abs}}$, computed on the subset of actionable objects whose ground-truth is \texttt{UNPLACED} (``when not to place''). All metrics are computed per scene and averaged uniformly (macro). When aggregating across profiles, we report the unweighted mean of profile scores so that each profile contributes equally.

\section{Results}\label{sec:results}

\subsection{Results on PARSEC}\label{sec:results_parsec}
\noindent
Table~\ref{tab:tab2} reports results on PARSEC, novel environment categories. \textbf{APOLLO} achieves the best average placement accuracy (\textbf{0.61}), improving over CSLM by \textbf{+0.02} (0.61 vs. 0.59), APRICOT-NI by \textbf{+0.08} (0.61 vs. 0.53), and TidyBot-Random by \textbf{+0.17} (0.61 vs. 0.44). The improvement is modest but stable across categories: APOLLO matches the best score on U-1D (0.56), matches CSLM on U-2D (0.57), and improves on Non-Uniform by \textbf{+0.02} (0.67 vs. 0.65). APOLLO uses the LLM for only a subset of decisions (routes 23.15\% of objects across 61.76\% of scenes), yielding this accuracy improvement while reducing LLM usage; we also complement these results with runtime measurements (PEM: 0.102 s/scene on CPU; CSLM: 8.401 s/scene).

As an ablation, \textbf{PEM} alone reaches \textbf{0.59}, matching CSLM on average (0.59) while remaining competitive across categories (U-1D: 0.56 vs. 0.54, Non-Uniform: 0.65 vs. 0.65), and conceding some performance on U-2D (0.54 vs. 0.57). This highlights that a learned personalized embedding model can recover much of the placement performance without LLM assistance, thus being the main contributor to APOLLO's performance.

\subsection{Results on APOR}\label{sec:results_profiles}

\noindent
Table~\ref{tab:tab3} compares APOLLO against CSLM and the PEM ablation on APOR. Overall, APOLLO achieves the highest mixed accuracy (\(\mathrm{pa}_{mix}=0.39\)) by selectively leveraging the LLM when PEM is uncertain, while preserving PEM’s strength on abstention (\(\mathrm{pa}_{abs}=0.47\)). In contrast, CSLM attains the strongest placement on average (\(\mathrm{pa}_{place}=0.30\)) but struggles to abstain (\(\mathrm{pa}_{abs}=0.22\)), yielding a weaker overall \(\mathrm{pa}_{mix}=0.25\). PEM exhibits the opposite behavior: it is highly reliable at abstention (\(\mathrm{pa}_{abs}=0.56\)) but weaker on pure placement (\(\mathrm{pa}_{place}=0.17\)), resulting in \(\mathrm{pa}_{mix}=0.38\). This asymmetry motivates the routing approach: APOLLO recovers part of CSLM’s placement advantage while keeping a substantial abstention capability.

At the profile level, the same pattern is visible. When abstention is dominant (e.g., Exclusionist), PEM already matches the best mixed score (\(\mathrm{pa}_{mix}=0.77\)), and APOLLO does not gain from escalation because the LLM provides little additional value for these instances. Conversely, profiles with moderate abstention rates benefit more from routing: APOLLO improves PEM’s placement without collapsing abstention, increasing \(\mathrm{pa}_{mix}\) on Frequency and Ladybug (\(+0.05\) each) and HighReach (\(+0.03\)). Importantly, APOLLO also exceeds CSLM on placement in some profiles (e.g., Frequency and Ladybug), suggesting that escalation can act as a complementary mechanism rather than a fallback.

Overall, Table~\ref{tab:tab3} provides proof-of-concept evidence that personalized rearrangement benefits from reasoning about both \emph{where to place} and \emph{when not to place}, with PEM and LLM-based reasoning exposing complementary strengths under the APOR benchmark.

\begin{table}[t]
\centering
\caption{
\textbf{Runtime efficiency.} Inference time comparison of APOLLO with baselines on the PARSEC and APOR datasets.}
\label{tab:efficiency_runtime}
\begin{tabular}{lccc}
\toprule
Model & PARSEC & APOR & Speedup \\
\midrule

CSLM & 4.850 & 20.273 & 1.0$\times$ / 1.0$\times$ \\
PEM & \textbf{0.00017} & \textbf{0.00061} & 28.5k$\times$ / 33.2k$\times$ \\
APOLLO & \underline{1.12} & \underline{11.56} & 4.3$\times$ / 1.8$\times$ \\
\bottomrule
\end{tabular}
\end{table}

\paragraph*{Runtime and routing efficiency.}
Beyond accuracy, APOLLO is motivated by reducing dependence on expensive LLM inference. Table~\ref{tab:efficiency_runtime} reports runtime per evaluated case and the corresponding speedup relative to full CSLM reasoning. PEM inference is several orders of magnitude faster than CSLM, and APOLLO translates this into an effective runtime reduction of about 4.3$\times$ on PARSEC and 1.8$\times$ on APOR. The smaller APOR speedup reflects its higher routing rate, with 57.0\% of actionable objects routed to the LLM compared with 23.1\% on PARSEC.

\begin{table}[h]
\centering
\footnotesize
\setlength{\tabcolsep}{5pt}
\renewcommand{\arraystretch}{1.05}
\caption{\textbf{APOR results by environment complexity.} Environments are grouped into terciles (Low/Mid/High) by the geometric complexity score. Each cell reports the triplet $\mathrm{pa}_{mix}\mid\mathrm{pa}_{place}\mid\mathrm{pa}_{abs}$.}
\begin{tabular}{lccc}
\toprule
\textbf{Model} & \textbf{Low} & \textbf{Mid} & \textbf{High} \\
\midrule
CSLM & 0.28 $|$ \textbf{0.32} $|$ 0.25 & 0.24 $|$ \textbf{0.29} $|$ 0.21 & 0.22 $|$ \textbf{0.27} $|$ 0.16 \\
PEM  & \underline{0.39} $|$ 0.18 $|$ \textbf{0.62} & \underline{0.37} $|$ 0.17 $|$ \textbf{0.59} & \underline{0.34} $|$ 0.15 $|$ \textbf{0.57} \\
APOLLO  & \textbf{0.41} $|$ \underline{0.27} $|$ \underline{0.51} & \textbf{0.39} $|$ \underline{0.24} $|$ \underline{0.52} & \textbf{0.33} $|$ \underline{0.24} $|$ \underline{0.38} \\
\bottomrule
\end{tabular}
\label{tab:tab4}
\end{table}

\subsubsection*{APOR: Environments complexity}\label{sec:results_envcomplexity}
\noindent
Table~\ref{tab:tab4} monitors how performance evolves on APOR as environments become more realistic, by grouping the \((u,e)\) environments into terciles of geometric complexity. The main takeaway is that increasing complexity consistently hurts all methods, suggesting that room-level environments increase the difficulty of personalized rearrangement. At the same time, each approach largely retains its core strength across terciles: CSLM remains strongest on placement (\(\mathrm{pa}_{place}: 0.32 \rightarrow 0.29 \rightarrow 0.27\)), while PEM remains strongest on abstention (\(\mathrm{pa}_{abs}: 0.62 \rightarrow 0.59 \rightarrow 0.57\)). Notably, CSLM’s abstention degrades more sharply than PEM’s placement (\(\mathrm{pa}_{abs}: 0.25 \rightarrow 0.21 \rightarrow 0.16\) vs. \(\mathrm{pa}_{place}: 0.18 \rightarrow 0.17 \rightarrow 0.15\)), reinforcing that abstention is a distinct challenge. APOLLO preserves the strengths of both models in the Low and Mid terciles (\(\mathrm{pa}_{mix}=0.41\) and \(0.39\)), but its abstention drops markedly in the High tercile (\(\mathrm{pa}_{abs}: 0.52 \rightarrow 0.38\)), indicating that the most complex environments remain challenging even with LLM assistance.

\subsubsection*{APOR: Corrupted partial scenes}\label{sec:results_partialerrors}
\noindent 
Table~\ref{tab:pa_mix_k5_k10_errors} isolates the impact of injecting errors in the in-scene context for partial scenes with \(k\in\{5,10\}\) observed objects. Overall, the effect on \(\mathrm{pa}_{mix}\) is modest, but it is consistently non-zero, supporting the motivation that assuming a perfectly reliable partial observation can slightly bias conclusions. For PEM, performance is essentially stable under noise (\(-0.003\) at \(k{=}5\), \(-0.001\) at \(k{=}10\)), suggesting that its context scoring is relatively robust to small perturbations. APOLLO inherits this robustness but still degrades mildly when the routed decision relies on a potentially corrupted context (\(-0.003\) at \(k{=}5\), \(-0.005\) at \(k{=}10\)). In contrast, CSLM shows the largest drop at \(k{=}10\) (\(-0.014\)), while the slight improvement at \(k{=}5\) (\(+0.003\)) likely reflects variance rather than a systematic benefit from noise. Taken together, these results indicate that APOLLO (and PEM) remain relatively stable when there are already partially arranged environments. Additionally, the context noise does not dominate performance for our model, but it can selectively affect LLM-based methods that rely heavily on in-scene evidence, especially as the partial scene becomes denser.

\begin{table}[t]
\centering
\footnotesize
\caption{\textbf{APOR results under corrupted partial scenes.} We compare $\mathrm{pa}_{mix}$ between clean partial scenes (\textit{correct}) and noisy partial scenes (\textit{errors}) for $k\in\{5,10\}$ placed-object modes (1 error for $k{=}5$, 2 errors for $k{=}10$).
}
\label{tab:pa_mix_k5_k10_errors}
\setlength{\tabcolsep}{4.5pt}
\renewcommand{\arraystretch}{1.05}
\begin{tabular}{lcc cc}
\toprule
\textbf{Model} &
\multicolumn{2}{c}{\textbf{$k{=}5$ (1 error)}} &
\multicolumn{2}{c}{\textbf{$k{=}10$ (2 errors)}} \\
\cmidrule(lr){2-3}\cmidrule(lr){4-5}
& \textbf{correct} & \textbf{errors} & \textbf{correct} & \textbf{errors} \\
\midrule
CSLM   & 0.251 & 0.254 & 0.243 & 0.229 \\
PEM    & \underline{0.296} & \underline{0.293} & \textbf{0.370} & \textbf{0.369} \\
APOLLO & \textbf{0.327} & \textbf{0.324} & \underline{0.359} & \underline{0.354} \\
\bottomrule
\end{tabular}
\end{table}


Overall, these results demonstrate that by using LLMs as assistance models with personalized embeddings, rather than zero-shot semantic reasoners (e.g., CSLM), APOLLO can learn personalized object placement preferences for simpler and complex settings across a variety of users. 

\section{Conclusion}
\label{sec:conclusion}

\noindent
In this paper, we address abstention-aware personalized object rearrangement and introduce APOLLO, a hybrid framework that combines a lightweight personalized embedding model with LLM assistance. APOLLO uses the local model for confident decisions and selectively invokes LLM reasoning for lower-confidence cases, reducing LLM usage while preserving complementary placement and abstention behavior. We also introduce APOR, a synthetic LLM-generated benchmark for studying room-level, multi-furniture environments, diverse organizational profiles, explicit abstention behavior, and noisy partial context. Experiments on PARSEC and APOR provide proof-of-concept evidence that separating \emph{where to place} from \emph{when not to place} is useful for personalized rearrangement, and that local personalized models and LLM-based reasoning can play complementary roles. However, APOR remains synthetic and is not a substitute for validation with real users or deployed robots. Future work will evaluate this framework in HRI studies with human participants and extend it to multi-user environments and long-term adaptation as user preferences evolve over time~\cite{ayub2024human}.


\bibliographystyle{IEEEtran}
\bibliography{Paper}

\end{document}